\documentclass{article}

\usepackage{iclr2027_conference,times}
\usepackage{wrapfig}
\usepackage{booktabs}
\usepackage{xcolor}
\usepackage{colortbl} % 彩色表格
\usepackage[most]{tcolorbox}
\usepackage{multirow}
\usepackage{subcaption}
\usepackage{lmodern}
\usepackage{fancyvrb}
\usepackage{fvextra}
\usepackage{titlesec}
\usepackage{parskip}
\tcbuselibrary{listings,theorems,breakable}
\newtcbtheorem[number within=section]{exmp}{Prompts}%
{breakable,colback=white!5!white,colframe=black!95!,fonttitle=\bfseries, left=.02in, right=.02in,bottom=.02in, top=.02in}{exmp}

\newtcbtheorem[number within=section]{case}{Examples}%
{breakable,colback=white!5!white,colframe=black!95!,fonttitle=\bfseries, left=.02in, right=.02in,bottom=.02in, top=.02in}{case}

\usepackage[utf8]{inputenc} % allow utf-8 input
\usepackage[T1]{fontenc}    % use 8-bit T1 fonts
\usepackage{hyperref}       % hyperlinks
\usepackage{url}            % simple URL typesetting
\usepackage{booktabs}       % professional-quality tables

\usepackage{siunitx}
\usepackage{nicefrac}       % compact symbols for 1/2, etc.
\usepackage{microtype}      % microtypography
\usepackage{xcolor}         % colors
\definecolor{purple1}{HTML}{e5e0f0}

\newcommand{\best}{\cellcolor[HTML]{e5e0f0}} % C8BFD9
\usepackage[utf8]{inputenc} % allow utf-8 input
\usepackage[T1]{fontenc}    % use 8-bit T1 fonts
\usepackage{amsfonts}       % blackboard math symbols
\usepackage{hyperref}       % hyperlinks
\usepackage{url}            % simple URL typesetting
\usepackage{booktabs}       % professional-quality tables
\usepackage{amsfonts}       % blackboard math symbols
\usepackage{nicefrac}       % compact symbols for 1/2, etc.
\usepackage{microtype}      % microtypography
\usepackage{xcolor}         % colors
\usepackage{amsmath}
\usepackage{amssymb}
\usepackage{mathtools}
\usepackage{amsthm}
\usepackage{caption}
\usepackage{graphicx}
\usepackage{multirow}
\usepackage{enumitem}
\usepackage{color}
\usepackage{xcolor}
\usepackage{colortbl}
\usepackage{pifont}
\usepackage{algorithm}
\usepackage{algorithmic}
\usepackage{ulem}
\usepackage{booktabs}
\usepackage{bbding}
\usepackage{wrapfig}
\definecolor{DarkGreen}{RGB}{1,100,32} 
\usepackage[capitalize,noabbrev]{cleveref}

\theoremstyle{plain}
\newtheorem{theorem}{Theorem}[section]

\theoremstyle{definition}

\newtheorem{assumption}[theorem]{Assumption}
\theoremstyle{remark}

\definecolor{titleblue}{HTML}{1F4E79}
\definecolor{softbg}{HTML}{F7FAFC}
\definecolor{lineblue}{HTML}{90CAF9}
\definecolor{taggray}{HTML}{4A5568}

\newtcolorbox{promptbox}[1][]{
  enhanced,
  breakable,
  colback=softbg,
  colframe=lineblue,
  boxrule=0.6pt,
  arc=2mm,
  left=2mm,
  right=2mm,
  top=1mm,
  bottom=1mm,
  #1
}

\DefineVerbatimEnvironment{PromptVerbatim}{Verbatim}{
  fontsize=\small,
  breaklines=true,
  breakanywhere=true
}

\title{DCRL: Decoupling and Coupling Reinforcement Learning via Policy-Reward Manifold Alignment}

\author{
\textbf{Henan Sun\textsuperscript{1}},
 \textbf{Zehua Li\textsuperscript{1}},
 \textbf{Haitao Hu\textsuperscript{1}},
 \textbf{Qifan Zhang\textsuperscript{1}},
 \textbf{Jianfeng Zhang\textsuperscript{3}},
 \textbf{Nuo Chen\textsuperscript{1,4}},
 \textbf{Jia Li \textsuperscript{1,2,*}},
\\
\\
 \textsuperscript{1}The Hong Kong University of Science and Technology (Guangzhou),\\
 \textsuperscript{2}The Hong Kong University of Science and Technology,\\
 \textsuperscript{3}Huawei Noah's Ark Lab,\\
 \textsuperscript{4}Tencent HY
}

\begin{document}

\iclrfinalcopy
\maketitle
%\pagestyle{fancy}
%\fancyhead[L]{Under review as a conference paper at ICLR 2027}

\begin{abstract}

Reinforcement learning (RL) has emerged as a key paradigm for improving the reasoning capabilities of large language models (LLMs). However, existing reward systems, such as rule-based and reward-model-based, often exhibit issues such as unstable optimization and reward hacking.
In this work, we revisit the general reasoning of LLMs from a geometric perspective, conceptualizing it as a coupled manifold composed of three interdependent sub-manifolds: logical deduction, evaluation, and representation. Based on this perspective, response generation in RL can be interpreted as a decoupling process from the evaluation manifold, while reward estimation corresponds to a decoupling process from the logical deduction manifold. The limitations of rule-based and reward-model RL systems can be geometrically interpreted as the mismatch of policy-reward manifolds during RL process.
To address the aforementioned misalignment, we propose Decoupling and Coupling Reinforcement Learning (DCRL) framework, which incorporates two key components: (1) a syllogistic logic-based prompt evolution mechanism that dynamically refines reward rubrics to enhance the expressiveness of the reward manifold; and (2) a policy–reward re-coupling mechanism that jointly updates the reward and policy models, ensuring consistent evaluation and mitigating manifold mismatch during training.
Theoretical analysis and extensive experiments across multiple reasoning domains demonstrate that DCRL consistently outperforms both rule-based and reward-model baselines. Notably, a Qwen3-4B model trained under DCRL surpasses a Qwen3-32B baseline and approaches the performance of a Qwen3-235B model, highlighting superior effectiveness and generalization in RL.
  
\end{abstract}

\section{Introduction}
\label{sec:Intro}
Reinforcement learning (RL) has demonstrated substantial performance improvements of large language models (LLMs)~\citep{shao2024deepseekmath,guo2025deepseekr1}. In this paradigm, the parameters of a policy model are iteratively optimized to maximize rewards provided by a reward system~\citep{stiennon2022learning,ouyang2022training,dong2024rlhf,kaufmann2025survey}, which may be defined through rule-based mechanisms or learned reward models~\citep{guo2025deepseekr1,dong2024rlhf}. The efficacy of such a reward system fundamentally depends on its ability to reliably and accurately evaluate the quality of model-generated responses within the context of specific queries~\citep{lambert2024rewardbench,liu2025compassverifier,zhang2025compassjudger2}.

Current reward systems in RL post-training are mainly divided into rule-based systems and reward model systems. Rule-based systems assign rewards through manually designed rules, heuristics, or programmatic verifiers, providing a simple and interpretable mechanism for supervising model behavior~\citep{guo2025deepseekr1,feng2026rlar}. However, because such rules are handcrafted, they often fail to accurately capture the true quality of model outputs, especially in complex or open-ended tasks, which can cause models to optimize for superficial rule satisfaction rather than genuine performance improvements~\citep{gunjal2025rubrics,zhang2025chasing,skalse2022defining,wen2024reward}. This limitation also makes rule-based systems vulnerable to reward hacking, where models exploit weaknesses in the reward design without improving actual task performance~\citep{chen2024odin,miao2024inform}. To overcome these issues, recent studies increasingly adopt reward model systems, which learn reward signals from preference data and provide more flexible supervision by implicitly modeling response quality and reasoning quality~\citep{stiennon2022learning,ouyang2022training,bai2022constitutional,lightman2023letsverify}. Although reward models have shown stronger empirical performance than rule-based methods~\citep{lambert2024rewardbench,liu2025skyworkrewardv2}, they still suffer from optimization instability, since iterative policy updates may exploit inaccuracies in the learned reward models, leading to reward over-optimization and degraded alignment with true task objectives~\citep{rafailov2024direct,skalse2022defining,wen2024reward}. Therefore, while reward model systems provide a more expressive alternative to rule-based rewards, ensuring their stability and alignment with true response quality remains a major challenge~\citep{lambert2024rewardbench,xu2025askstrongjudge}.

\begin{figure*}[t]
	\centering
  \includegraphics[width=0.6\textwidth]{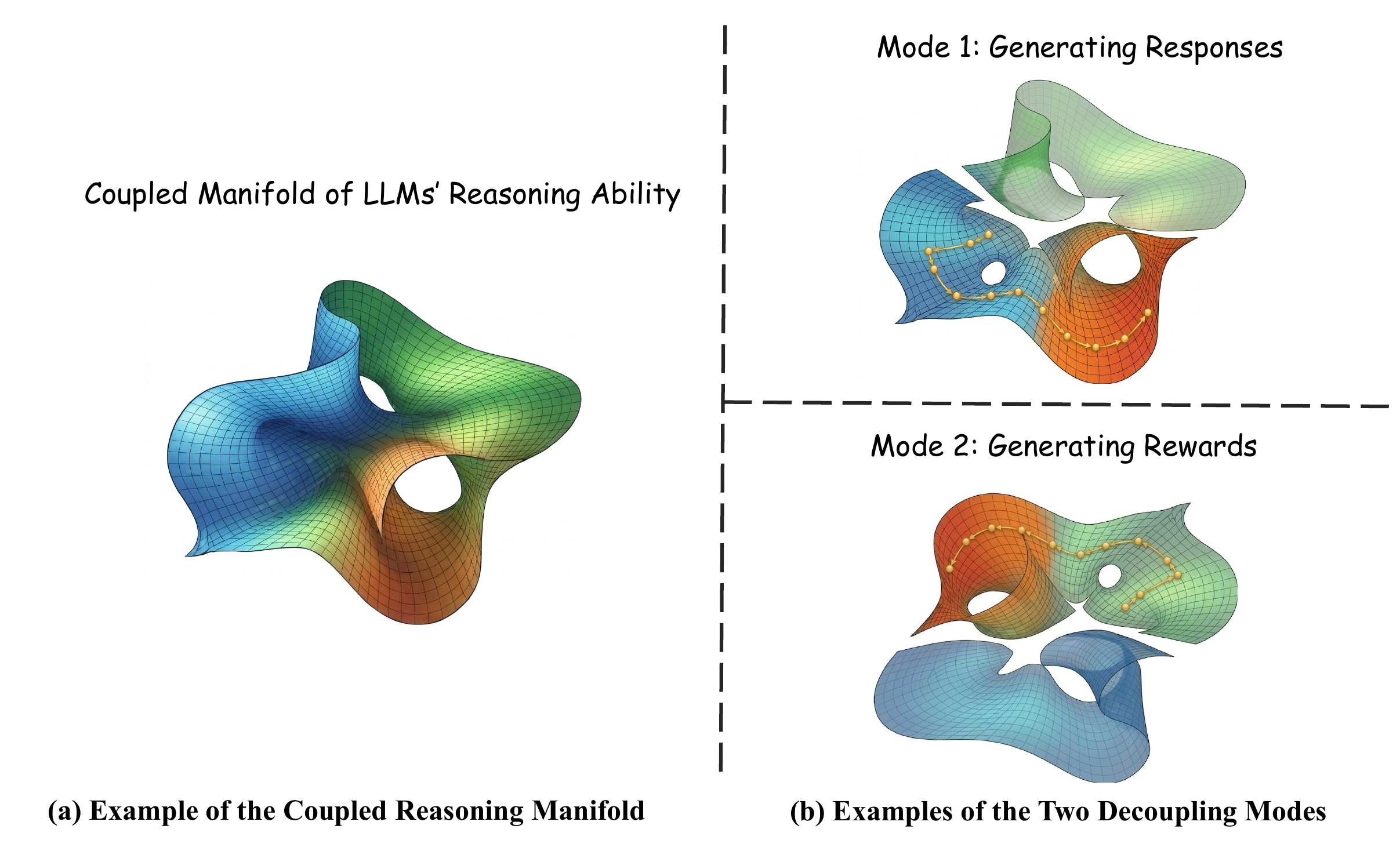}
  \caption{
    Illustration of coupled reasoning manifold and two decoupling modes.
}
\label{fig: intro}
\end{figure*}

In contrast to prior studies, we examine reward systems through a geometric lens on the nature of reasoning, conceptualizing it as a collection of coupled manifolds embedded in a high-dimensional space, each associated with distinct reasoning patterns, skills, or strategies (see Fig.~\ref{fig: intro}), inspired by recent progress from the manifold theory~\citep{brahma2015deep, magai2022topology}. Specifically, pretraining data induces a high-dimensional data manifold $\mathcal{M}$, within which heterogeneous reasoning sub-manifolds—namely, the logical deduction sub-manifold $\mathcal{M}_L$ (illustrated as the blue surface in Fig.~\ref{fig: intro} (a)), the evaluation sub-manifold $\mathcal{M}_E$ (green surface), and the representation sub-manifold $\mathcal{M}_R$ (orange surface)—are organized into a stratified structure, formally expressed as $\mathcal{M} = \mathcal{M}_L \cup \mathcal{M}_E \cup \mathcal{M}_R$.
From this perspective, reasoning can be formulated as the interaction among three interdependent sub-manifolds: (i) the logical deduction sub-manifold $\mathcal{M}_L$, which is coupled with the model’s policy distribution over $\mathcal{M}$; (ii) the evaluation sub-manifold $\mathcal{M}_E$, encoding the implicit judgment functions underlying logical inference; and (iii) the representation sub-manifold $\mathcal{M}_R$, in which both model responses and evaluation criteria are explicitly instantiated in the text space. As illustrated in Fig.~\ref{fig: intro} (b), the generation of responses conditioned on a query can be interpreted as a geometric process in which the reasoning manifold effectively decouples from the evaluation sub-manifold, formalized as $\mathcal{M} \perp \mathcal{M}_E$. Conversely, when the model assigns rewards to assess response quality, this corresponds to a geometric process in which the reasoning manifold decouples from the logical deduction sub-manifold, i.e., $\mathcal{M} \perp \mathcal{M}_L$.
From the geometric standpoint outlined above, existing approaches exhibit several inherent limitations when analyzed through the framework of manifold interactions:

\textit{Limitation 1}: Rule-based reward systems largely disregard the evaluation sub-manifold embedded within the overall reasoning manifold $\mathcal{M}$ of the model, thereby limiting their capacity to capture nuanced judgment processes. Consequently, they are particularly susceptible to critical challenges, including poor adaptability to open-ended settings and vulnerability to reward hacking~\citep{skalse2022defining,gunjal2025rubrics,wen2024reward}. 

\textit{Limitation 2}: Reward model systems typically rely on static reward models whose evaluation sub-manifold does not adapt in tandem with the evolving reasoning manifold of the policy model~\citep{uo2024selfrewarding,wu2025rmrouting,xu2025askstrongjudge,feng2026rlar}. This mismatch can result in several critical challenges, including unstable optimization dynamics, misalignment between reward signals and true task performance, and increased susceptibility to reward hacking~\citep{liu2025deepseekgrm,wu2026fastslow,xue2026reasononly}.

Observing the aforementioned limitations, we derive two key motivations as following. \textbf{(M1)}: \textbf{The reward rubric should evolve dynamically to enhance the expressiveness of reward evaluation.} Since the reasoning behavior of the policy model changes continuously during training, a static reward rubric may fail to capture the evolving evaluation criteria required for accurate assessment. By adaptively refining the reward rubric according to the current training state, the reward model can better externalize and calibrate its implicit evaluation standards, thereby producing more precise and reliable reward signals. \textbf{(M2)}: \textbf{The reward model should evolve jointly with the policy model to preserve evaluation consistency during reinforcement learning.} As the reasoning capability of the policy model improves, a static reward model may gradually become misaligned with the policy model’s evolving output distribution, resulting in inaccurate reward estimation. To maintain reliable evaluation, improvements in policy reasoning should be coupled back to the reward model, thereby reducing the mismatch between policy and reward models and ensuring consistent reward supervision throughout training.

Based on the motvations above, we propose a novel framework termed \textit{decouple-and-couple} reward modeling system, along with a corresponding reinforcement learning paradigm, \textbf{D}ecouple and \textbf{C}ouple \textbf{RL} (DCRL), inspired by the following motivations.
 The proposed approach first decouples the evaluation and representation sub-manifolds from the originally entangled manifold induced by the model, then independently optimizes these sub-manifolds via our proposed syllogistic logic prompt-evolving mechanism, and finally re-integrates them through RL-based coupling. This structured decoupling and subsequent coupling process enables more stable optimization dynamics and yields consistent performance improvements.

\textbf{Our Contributions.} 
(1) \textit{\underline{New Perspective.}} 
To the best of our knowledge, this work is the first to investigate reward modeling systems through the geometric nature on reasoning, by conceptualizing the process as the interaction of three coupled sub-manifolds: logical deduction, evaluation, and representation. This formulation provides a principled and theoretically grounded foundation for the design of our reinforcement learning framework.
(2) \textit{\underline{Novel RL System.}} 
We propose a novel paradigm termed DCRL, which first disentangles the evaluation and representation sub-manifolds from the originally coupled manifold induced by the model. These sub-manifolds are then independently optimized via a syllogistic logic–based prompt-evolving mechanism, and subsequently re-integrated through reinforcement learning–based coupling.
(3) \textit{\underline{Strong Empirical Performance.}}
Extensive experiments across diverse reasoning domains, including mathematics, code, and commonsense, demonstrate that DCRL consistently outperforms both rule-based and static reward model baselines. Notably, a Qwen3-4B model trained with DCRL surpasses a Qwen3-32B counterpart and approaches the performance of a Qwen3-235B model, highlighting the effectiveness and generalization of the proposed DCRL paradigm for general reasoning.
\section{Related Work}
\label{sec:prelim_related}

Recent advances in reward modeling for long-form reasoning post-training can be broadly categorized into \textbf{rule-based systems} and \textbf{reward model systems}.

\textbf{Rule-based reward systems} assign reward signals according to predefined heuristic rules or task-specific verifiers, offering a simple and reliable mechanism for reinforcement learning without requiring separately trained reward models. This paradigm has been widely adopted in RL, especially in domains where objective correctness can be automatically verified. For instance, DeepSeekMath~\citep{sha2024deepseekmath} employs rule-based rewards derived from symbolic answer matching to reinforce mathematical reasonin. Similarly, Math-Shepherd~\citep{wang2024mathshepherd} introduces automatically verifiable step-level supervision signals, where rewards are assigned according to rule-based verification of intermediate reasoning correctness. In code generation, CodeRL~\citep{le2022coderl} leverages unit test execution results as rule-based rewards, where generated programs are rewarded based on whether they pass predefined tests. Extending this paradigm, RLEF~\citep{dong2023rlef} uses compiler and execution feedback as deterministic rule-based rewards for reinforcement learning from execution feedback, significantly improving program synthesis performance.

\textbf{Reward model systems} assign reward signals to model outputs through learned reward models and have attracted increasing attention in reinforcement learning research due to their conceptual simplicity and computational efficiency. This line of work includes outcome reward models, preference models, and process reward models. For example, Skywork-Reward-V2~\citep{liu2025skyworkrewardv2} improves reward modeling through large-scale human-AI collaborative preference data curation, establishing strong open-source reward model baselines. ReasonFlux-PRM~\citep{zou2025reasonflux} extends process reward modeling to long chain-of-thought reasoning by introducing trajectory-aware reward supervision. CodePRM~\citep{li2025codeprm} incorporates execution feedback to improve process-level evaluation for code generation tasks. CompassVerifier~\citep{liu2025compassverifier} further enhance scalar evaluation quality by training specialized verifiers for reasoning tasks with objective correctness signals. DeepSeek-GRM~\citep{liu2025deepseekgrm} proposes a generalist generative reward model trained via Self-Principled Critique Tuning, enabling adaptive reward generation during evaluation. Owing to their efficiency and ease of integration, reward model systems serve as a practical foundation for reward-guided policy optimization.

\section{Methodology}
\label{sec: Metho}

\subsection{Motivation}
\label{sec: Motivation}
As illustrated in Section~\ref{sec:Intro}, existing reward systems in RL suffer from the following limitations: (1) rule-based reward systems largely ignore the evaluation sub-manifold coupled within the overall reasoning manifold of the model, leading to poor adaptability to open-ended settings and vulnerability to reward hacking. (2) reward model systems suffer from the mismatch between the evolving reasoning manifold of the policy model and the static evaluation sub-manifold of the reward model, leading to the unstable optimization and reward hacking problems.

Inspired by the recent progress on the manifold theory~\citep{brahma2015deep, magai2022topology}, we conceptualize the abstract reasoning ability of LRMs as the coupled manifold illustrated as Fig.~\ref{fig: intro} in Section~\ref{sec:Intro}. In this view, different tasks in RL correspond to distinct patterns of manifold coupling rather than a single unified reasoning process. Specifically, when an LLM generates an answer, it primarily operates through the coupling between the logical deduction sub-manifold and the representation sub-manifold: the model performs step-by-step reasoning in its latent space and subsequently maps this reasoning trajectory into textual form. In contrast, when an LLM evaluates or scores an answer, the process relies on the coupling between the evaluation sub-manifold and the representation sub-manifold, where internal judgment criteria are expressed through natural language outputs.

This perspective reveals a fundamental mismatch in standard RL frameworks for LLM alignment. When the reward model or reward rubric is held static, its evaluation–representation sub-manifold may gradually become misaligned with the evolving logical deduction–representation sub-manifold of the policy model, leading to degraded reward fidelity and suboptimal optimization. Recognizing that the reward model itself constitutes a sub-manifold of the overall reasoning manifold, we argue that improvements in reasoning induced by RL should naturally extend to the evaluation sub-manifold. This suggests a simple yet effective strategy: directly updating the reward model using the improved policy model, thereby inheriting enhanced reasoning and evaluation capabilities.
Furthermore, since the reward signal is ultimately produced through the coupling between evaluation and representation sub-manifolds, improving the expressiveness of the representation sub-manifold can further enhance reward quality. To this end, we propose to explicitly evolve the representation component of the reward model via syllogistic logic prompt-evolving, enabling the model to externalize and refine its implicit evaluation criteria. Together, these insights motivate the proposed DCRL in which both policy and reward models co-evolve through the dynamic coupling of reasoning sub-manifolds, leading to more accurate evaluation and more effective alignment.

\subsection{DCRL Framework}
\label{sec: framework}

\begin{algorithm}[t]
\caption{Decoupling and Coupling Reinforcement Learning (DCRL)}
\label{alg: dcrl}
\begin{algorithmic}[1]

\STATE \textbf{Input:} Pretrained policy model
$\pi_{\theta_0}$, initial reward model $r_{\phi_0}$ with the same parameter weights of the policy model, 
axiomatic prompts $\mathcal{A}=\{a_i\}_{i=1}^{k_1}$, 
theorem prompts $\mathcal{T}=\{t_j\}_{j=1}^{k_2}$, 
corollary prompts $\mathcal{C}_0=\{c_l\}_{l=1}^{k_3}$

\FOR{epoch $e = 0,1,2,\dots,E$}

    \STATE \textbf{// Step 1: Sample trajectories}
    \STATE Sample $x \sim \pi_{\theta_e}(x)$

    \STATE \textbf{// Step 2: Reward computation via coupled manifolds}
    \STATE Construct rubric prompt:

    \begin{equation}
    \label{eq: rubric}
    \mathcal{R}_e = \mathcal{A} \cup \mathcal{T} \cup \mathcal{C}_e
    \end{equation}
    
    \STATE Compute reward:
    \begin{equation}
    \label{eq: compute_reward}
    r_{\phi_e}(x) = f_{\phi_e}(x; \mathcal{R}_e)
    \end{equation}

    \STATE \textbf{// Step 3: Policy optimization}
    \STATE Compute advantage:
    \begin{equation}
    \label{eq: compute_advantage}
    A_e(x) = r_{\phi_e}(x) - b_e
    \end{equation}
    
    \STATE Update policy:
    \begin{equation}
    \label{eq: update_policy}
    \pi_{\theta_{e+1}}(x) \propto 
    \pi_{\theta_e}(x)\exp\big(\eta A_e(x)\big)    
    \end{equation}

    \STATE \textbf{// Step 4: Corollary prompt evolution}
    \STATE Update $\mathcal{C}_e \rightarrow \mathcal{C}_{e+1}$ via:
    \begin{equation}
    \label{eq: update_corollary}
    \begin{aligned}
    \mathcal{C}_{e+1} &= g_\phi\bigl(\mathcal C_t,\bar A_t,\mathrm{Acc}_t\bigr) \\
    &\approx \arg\max_{x \sim \pi_{\theta_e},\mathcal{C}}
    (1-\lambda_{acc})\mathbb{E}
    \big[ A_e(x)\big]+ \lambda_{acc}f_{\theta_e}(x; \mathcal{A}, \mathcal{T}, \mathcal{C}_e)
    \end{aligned}
    \end{equation}
    
    \STATE \textbf{// Step 5: Domain-level theorem update}
    \IF{reasoning domain shift detected}
        \STATE Update $\mathcal{T}$:
        \begin{equation}
        \label{eq: update_theorem}
        \begin{aligned}
    \mathcal{T} &= g_\phi\bigl(\mathcal T,\bar A_t,\mathrm{Acc}_t\bigr) \\
    &\approx \arg\max_{x \sim \pi_{\theta_e},\mathcal{T}}
    (1-\lambda_{acc})\mathbb{E}
    \big[ A_e(x)\big]+ \lambda_{acc}f_{\theta_e}(x; \mathcal{A}, \mathcal{T}, \mathcal{C}_e)
    \end{aligned}
    \end{equation}
    \ENDIF

    \STATE \textbf{// Step 6: Parameter coupling}
    \STATE Update reward model:
    \begin{equation}
    \label{eq: policy_updation}
    \phi_{e+1} \leftarrow \theta_{e+1}
    \end{equation}

\ENDFOR

\STATE \textbf{Output:} Final policy $\pi_{\theta_E}$

\end{algorithmic}
\end{algorithm}

Building upon the aforementioned motivation, the overall pipeline of the proposed DCRL is illustrated in pseudo code Algorithm~\ref{alg: dcrl}, which explicitly models and leverages the interaction between different reasoning sub-manifolds. Concretely, DCRL adopts a standard reinforcement learning backbone (e.g., GRPO~\citep{sha2024deepseekmath}) for policy optimization, while introducing a syllogistic logic–based prompt-evolving mechanism to dynamically refine the reward model. As shown in Eq.~(\ref{eq: rubric}), we decompose the reward model’s rubric prompts into three hierarchical components: (i) a set of $k_1$ axiomatic prompts $\mathcal{A}$, which encode fundamental evaluation principles that are universally applicable across reasoning domains (e.g., mathematical, coding, and commonsense reasoning) and remain fixed throughout training; (ii) a set of $k_2$ theorem prompts $\mathcal{T}$, which capture domain-specific evaluation criteria that are invariant within a given domain but can be adapted when transferring across domains; and (iii) a set of $k_3$ corollary prompts at 0-th step $\mathcal{C}_0$, which are dynamically updated during training based on the current behavior of the policy model, enabling fine-grained and adaptive evaluation. More specifically, following the common standard of RL backbones (e.g., GRPO), we calculate the advantage and update the policy logits illustrated as Eq.~(\ref{eq: compute_advantage}) and Eq.~(\ref{eq: update_policy}). The corollary prompt at $e$-th step $\mathcal{C}_e$ is updated according to Eq.~(\ref{eq: update_corollary}) to maximize the expectation of the corollary prompt conditioned on the advantage and the current training accuracy, where $g_\phi\bigl(\mathcal C_t,\bar A_t,\mathrm{Acc}_t\bigr)$ is the prompt-updating operator and $f_{\theta_e}(x; \mathcal{A}, \mathcal{T}, \mathcal{C}_e)$ is the accuracy calculation function. Similarly, if the reasoning domain is shift, the theorem prompt $\mathcal{T}$ is updated according to Eq.~(\ref{eq: update_theorem}). This hierarchical prompt evolution mechanism is designed to enhance the representation sub-manifold of the reward model, thereby improving the fidelity of the evaluation–representation coupling and yielding more accurate reward signals for the policy model.

In addition, to mitigate the structural mismatch between the reward model and the evolving policy model, we introduce a parameter synchronization strategy: at the end of each training epoch, the parameters of the reward model are directly replaced by those of the updated policy model, as illustrated in Eq.~(\ref{eq: policy_updation}). This operation is motivated by the observation that the reward model, as a sub-manifold of the overall reasoning manifold, should co-evolve with the policy model to maintain alignment between the evaluation–representation coupling and the logical deduction–representation coupling. By combining hierarchical prompt evolution with periodic parameter coupling, DCRL establishes a unified training paradigm in which the policy and reward models are alternately decoupled for specialized optimization and re-coupled for global consistency, leading to progressively improved reasoning performance and evaluation accuracy.

\subsection{Theoretical Analysis}
\label{sec: theo-analysis}

In this section, we provide the theoretical foundations of the proposed DCRL framework.
Our analysis focuses on the alignment dynamics between the reasoning manifold of the policy model and that of the reward model.
Specifically, we show that under the proposed reward-policy co-evolution mechanism, the discrepancy between the two reasoning manifolds remains bounded throughout training.
This bounded-gap property guarantees that the reward model remains aligned with the continually evolving policy model, thereby improving the stability and reliability of reward estimation.
The detailed proof is provided in Appendix~\ref{appendix: theo_proof}.

To formalize this result, we first introduce two assumptions.

\begin{assumption}[Prompt Smoothness]
\label{assump:smooth}
Let \(T_{\mathcal C}\) denote the prompt-induced transformation on the reward reasoning distribution under corollary prompts \(\mathcal C\).
There exists a constant \(L>0\) such that for any two prompt states \(\mathcal C\) and \(\mathcal C'\),
\begin{equation}
D_{\mathrm{KL}}
\bigl(
pT_{\mathcal C'} \,\|\, pT_{\mathcal C}
\bigr)
\le
L\|\mathcal C'-\mathcal C\|,
\label{eq:smoothness}
\end{equation}
where \(D_{\mathrm{KL}}(\cdot\|\cdot)\) denotes the Kullback-Leibler divergence.
This assumption implies that small prompt changes only induce small perturbations in the reward reasoning distribution.
\end{assumption}

\begin{theorem}
\label{theo: bounded}
Let the corollary prompt evolution operator be
\begin{equation}
\mathcal C_{t+1}
=
g_\phi\bigl(\mathcal C_t,\bar A_t,\mathrm{Acc}_t\bigr),
\label{eq:operator_update}
\end{equation}
where: (1) \(\bar A_t=\mathbb E_{x\sim\pi_{\theta_t}}[A_t(x)]\) is the average advantage at epoch \(t\), (2) \(\mathrm{Acc}_t\) is the training accuracy at epoch \(t\), (3) \(g_\phi\) is the prompt rewriting operator implemented by the reward model.

Then the prompt evolution step is bounded by a constant \(\delta>0\):
\begin{equation}
\|
\mathcal C_{t+1}-\mathcal C_t
\|
\le
\delta.
\label{eq:bounded_update}
\end{equation}
\end{theorem}

Based on the assumption and the theorem above, the discrepancy between the policy reasoning manifold and the reward reasoning manifold is bounded as follows.

\begin{theorem}
\label{theo: gap}
Under Assumptions~\ref{assump:smooth} and Theorem~\ref{theo: bounded}, the reasoning-manifold gap between the policy model and reward model satisfies
\begin{equation}
G_t
\le
2L\delta,
\qquad
\forall t,
\label{eq:main_gap}
\end{equation}
where \(G_t\) denotes the symmetric KL divergence between the policy reasoning distribution and the reward reasoning distribution at epoch \(t\).
\end{theorem}

Theorem~\ref{theo: gap} indicates that the discrepancy between the reward reasoning manifold and the policy reasoning manifold remains uniformly bounded throughout training.
Therefore, the reward model can continuously track the evolution of the policy reasoning process, ensuring stable reward evaluation and preventing manifold drift between the two models.
\section{Experiments}
\label{sec: experiments}
In this section, we present comprehensive experiments evaluating the proposed DCRL framework under diverse reasoning benchmark, including math, code and commen sense reasoning domains. The objective of these experiments is to address the following research questions: \textbf{Q1}: How does the proposed DCRL framework perform compared to other RL baselines? \textbf{Q2}: To what extent do the reasoning manifolds of policy and reward models align during the decoupling and coupling process? \textbf{Q3}: How about the training stability of the proposed DCRL compared to other RL paradigms? Due to page limit, we put part of experiment results and detailed analysis in Appendix~\ref{appen: pre-processing}.

\subsection{Experiment Setup}
\label{sec: experiment_setup}
\textbf{Benchmarks.} We evaluate DCRL on 10 benchmarks across three reasoning domains, as summarized in Table~\ref{tab:dataset-summary}. Our evaluation follows a two-stage protocol. Firstly, we conduct in-distribution evaluation on six source benchmarks with training splits: GSM8K and OlyBench for math, CodeContest and LiveCodeBench for code, and BBEH and MMLU-Pro for commonsense. Following prior settings~\citep{zou2025reasonflux, liu2025deepseekgrm}, each source benchmark is split into training and test sets with a 9:1 ratio for RL training and in-distribution evaluation. 
To further assess the generalization capability of DCRL, we introduce out-of-distribution (OOD) evaluation on evaluation-only benchmarks. For each domain, we select one trained policy for OOD testing: the policy trained on OlyBench is evaluated on AIME24 and AIME25, the policy trained in CodeContest on CRUX, and the policy trained in BBEH on GPQA. Evaluation is based on numeric match for math, unit test execution for code, and final answer exact match for commonsense. Additional details are provided in Appendix~\ref{appen: dataset_pre_processing}.

\textbf{Models \& Baselines.} In our experiments, Qwen3-4B is adopted as the policy model and trained under the DCRL framework, denoted as DCRL-4B (ours) in Table~\ref{tab:reward_system_comparison}. We further include Qwen3-4B, Qwen3-32B, and Qwen3-235B~\citep{yang2025qwen3} as zero-shot reference baselines to contextualize the upper-bound performance of DCRL-4B. To systematically evaluate the impact of different reward paradigms in RL, we further incorporate a rule-based reward system, where rewards are defined via handcrafted heuristics, as well as a static LLM-based reward generator that uses Qwen3-4B without rubric-based prompting. In addition, we include three strong and representative baselines from distinct research directions: ReasonFlux-7B~\citep{zou2025reasonflux}, DeepSeek-GRM-16B~\citep{liu2025deepseekgrm}, and Skywork-Reward-V2~\citep{liu2025skyworkrewardv2}—given their demonstrated effectiveness in general reasoning RL.

\textbf{Evaluation Metrics.} We report pass@5 for all tasks. Specifically, mathematical reasoning is evaluated by numeric match, code generation by unit test execution, and commonsense reasoning by final-answer exact match. To quantify the alignment between policy-reward reasoning manifolds during training, we employ Linear CKA~\citep{alvarez2022gaussian} and Procrustes Distance~\citep{kendall1984shape} as the metrics, where higher Linear CKA and lower Procrustes Distance corresponds to more aligned reasoning manifolds of policy and reward models. Detailed explanation about the metrics is illustrated in Appendix~\ref{appen: experiment_pre_processing}.

\textbf{Experiment Environment.} For reproducibility, we report the hardware and software configurations used in our experiments. All experiments were conducted on a server equipped with an Intel(R) Xeon(R) Gold 6240 CPU @ 2.60GHz and 8 NVIDIA A100 GPUs with 80GB memory, using CUDA 12.8.

\begin{table}[t]
\centering
\small
\setlength{\tabcolsep}{6pt}
\renewcommand{\arraystretch}{1.1}
\caption{Summary of dataset statistics.}
\resizebox{\textwidth}{!}{
\begin{tabular}{c c c c c c}
\toprule
\textbf{Domain} & \textbf{Dataset} & \textbf{Training Set Size} & \textbf{Testing Set Size} & \textbf{OOD} & \textbf{Evaluation Criteria} \\
\midrule

\multirow{4}{*}{Math}
& AIME24        & --     & 30    & \checkmark & Numeric Match \\
& AIME25        & --     & 30    & \checkmark & Numeric Match \\
& GSM8K         & 7,473  & 1,319 & $\times$  & Numeric Match \\
& OlyBench      & 1,913  & 213  & $\times$  & Numeric Match \\
\midrule

\multirow{3}{*}{Code}
& CodeContest   & 10,521 & 1,169 & $\times$ & Unit Test Execution \\
& L.C.Bench       & 656    & 73   & $\times$ & Unit Test Execution \\
& CRUX   & -- & 1,598 & \checkmark & Unit Test Execution \\
\midrule

\multirow{3}{*}{Commonsense}
& BBEH          & 2,499  & 279  & $\times$ & Final Answer Exact Match \\
& MMLU-Pro      & 10,891 & 1,211 & $\times$ & Final Answer Exact Match \\
& GPQA          & --    & 198   & \checkmark & Final Answer Exact Match \\
\bottomrule
\end{tabular}
}
\label{tab:dataset-summary}
\end{table}

\subsection{Performance Comparison}
\label{sec: performance_comparison}
To address \textbf{Q1}, we conduct a comprehensive empirical evaluation against a diverse set of baselines across multiple benchmarks spanning various reasoning domains, as reported in Table~\ref{tab:reward_system_comparison}. Based on the experimental results, we draw the following conclusions.
(\textbf{C1}): \textbf{The proposed DCRL-4B surpasses both Qwen3-4B and Qwen3-32B, and approaches the performance of Qwen3-235B, demonstrating the effectiveness of DCRL.} As shown in Table~\ref{tab:reward_system_comparison}, DCRL-4B consistently outperforms Qwen3-4B and Qwen3-32B across all evaluated benchmarks and remains competitive with Qwen3-235B. This result highlights the substantial improvement in general reasoning capability achieved by enhancing Qwen3-4B under the DCRL framework.
(\textbf{C2}): \textbf{DCRL-4B also outperforms counterparts based on rule-based and static reward model paradigms, validating the importance of mitigating the misalignment between policy and reward reasoning manifolds.} As discussed in Section~\ref{sec:Intro}, both rule-based and static reward model approaches fail to adequately address the discrepancy between policy and reward reasoning manifolds during training, often resulting in issues such as unstable optimization and reward hacking. In contrast, Theorem~\ref{theo: gap} provides theoretical support that DCRL effectively aligns these manifolds, thereby leading to improved performance across multiple benchmarks shown in Table~\ref{tab:reward_system_comparison}.
\begin{table*}[t]
\centering
\caption{Performance comparison of SOTA baselines and the proposed DCRL across Math, Code and Common sense benchmarks.}
\label{tab:reward_system_comparison}
\renewcommand{\arraystretch}{1.1}
\setlength{\tabcolsep}{4pt}
\resizebox{\textwidth}{!}{%
\begin{tabular}{c|cccc|ccc|ccc}
\toprule
\multirow{2}{*}{Baselines} 
& \multicolumn{4}{c|}{Math} 
& \multicolumn{3}{c|}{Code} 
& \multicolumn{3}{c}{Common Sense} \\
\cmidrule(lr){2-5} \cmidrule(lr){6-8} \cmidrule(lr){9-11}
& GSM8K & AIME24 & AIME25 & OlyBench 
& CodeContest & CRUX & L.C.Bench 
& BBEH & MMLU-Pro & GPQA \\
\midrule
Qwen3-4B
& 92.58 & 25.0 & 16.67 & 10.73
& 17.28 & 58.52 & 11.74
& 8.95 & 49.46 & 48.22 \\

Qwen3-32B    
& 93.9 & 26.7 & 20.0 & 20.2 
& 19.0 & 64.90 & 16.6 
& 10.75 & 58.5 & 61.92 \\ 

Qwen3-235B-A22B 
& \best{\textbf{96.74}} & \best{\textbf{44.18}} & \best{\textbf{40.67}} & \best{\textbf{54.08}} 
& \underline{35.79} & \underline{78.94} & \best{\textbf{42.58}} 
& \best{\textbf{40.75}} & \best{\textbf{71.35}} & 67.01 \\
\midrule
Rule-based   
& 94.81 & 26.67 & 20.0 & 13.88 
& 26.18 & 59.51 & 22.14 
& 31.09 & 60.06 & 60.75 \\

Static Reward Model   
& 94.54 & 33.33 & 30.0 & 14.27 
& 32.64 & 68.58 & 24.37 
& 32.32 & 60.42 & 64.75 \\

ReasonFlux-7B 
& 95.43 & 30.0 & 33.33 & 10.90 
& 16.00 & 69.92 & 27.44 
& 33.42 & 66.21 & \underline{71.72} \\

Deepseek-GRM-16B     
& 92.80 & 33.33 & 26.67 & 20.19 
& 13.11 & 59.01 & 23.42 
& 16.49 & 52.35 & 66.70 \\

Skywork-reward-v2   
& 93.63 & 33.33 & 26.67 & 16.43
& 16.51 & 72.66 & \underline{29.89}
& 11.11 & 50.25 & 62.31 \\

DCRL-4B (ours)  
& \underline{95.59} & \underline{40.0} & \underline{36.67} & \underline{21.12} 
& \best{\textbf{36.84}} & \best{\textbf{79.12}} & 26.95 
& \underline{35.13} & \underline{71.20} & \best{\textbf{77.27}} \\
\bottomrule
\end{tabular}
}
\end{table*}
%% 结果高亮用这个\best{\textbf{...}}

% Qwen3-235B-A22B-Thinking-2507 
% & \best{\textbf{95.68}} & 16.67 & 10.00 & 19.72 
% & 0.00 & -- & -- 
% & -- & 67.22 & 73.10 \\
% \midrule
% Rule-based   
% & 96.81 & 0 & 0 & 13.88 
% & 0 & 0 & 0 
% & 0 & 0 & 0 \\

\subsection{Manifolds Alignment Analysis}
\label{sec: manifolds_alignment}
\begin{figure*}[!htbp]
	\centering
  \includegraphics[width=\textwidth]{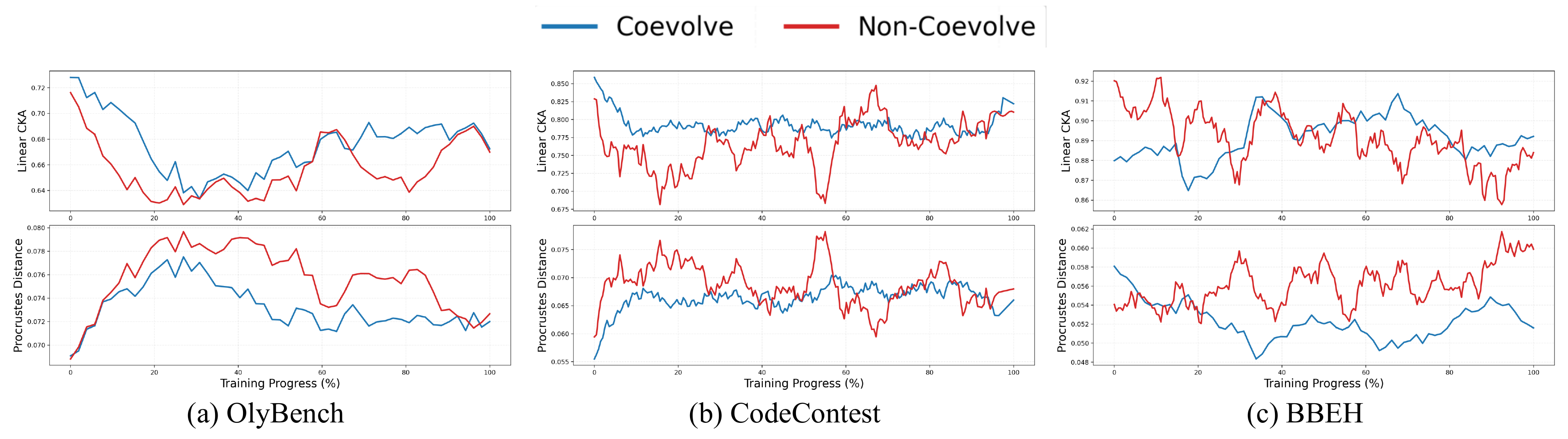}
  \caption{
    The policy-reward reasoning manifold alignment under DCRL and static reward system settings.
}
\label{fig: q2_alignment}
\end{figure*}

To address \textbf{Q2}, we compare models trained with a static reward system (non-coevolve) against those optimized under the proposed DCRL framework (coevolve) across three representative benchmarks spanning diverse reasoning domains: OlyBench for mathematical reasoning, CodeContest for code reasoning, and BBEH for commonsense reasoning, as shown in Fig.~\ref{fig: q2_alignment}. It can be observed that the coevolve setting consistently demonstrates stronger policy-reward manifold alignment than the non-coevolve baseline across all three benchmarks. Specifically, models trained under the proposed DCRL framework achieve generally higher Linear CKA values and lower Procrustes Distance, indicating improved structural similarity and reduced geometric discrepancy between the policy and reward reasoning manifolds. These results suggest that, under the DCRL framework, the reward model is able to adapt more effectively to the evolving reasoning manifold of the policy model during training. This empirical observation is consistent with the theoretical result established in Theorem~\ref{theo: gap}, which shows that the divergence between the policy reasoning distribution and the reward reasoning distribution at each training step is bounded by a constant. Such bounded divergence enables the reward manifold to evolve in coordination with the policy manifold, allowing the reward model to provide dynamically appropriate supervision signals throughout training. As a result, the policy model receives more accurate and adaptive reward guidance, which ultimately contributes to improved training effectiveness and enhanced reasoning performance.

% \begin{figure}[!htbp]
%     \centering
%     \begin{subfigure}[b]{0.33\textwidth}
%         \centering
%         \includegraphics[width=\textwidth]{q2_olybench_alignment.png}
%         \caption{OlyBench}
%         \label{fig: olybench_alignment}
%     \end{subfigure}
%     %\hfill
%     \begin{subfigure}[b]{0.33\textwidth}
%         \centering
%         \includegraphics[width=\textwidth]{q2_codecontest_alignment.png}
%         \caption{CodeContest}
%         \label{fig: codecontest_alignment}
%     \end{subfigure}
%     %\hfill
%     \begin{subfigure}[b]{0.33\textwidth}
%         \centering
%         \includegraphics[width=\textwidth]{q2_bbeh_alignment.png}
%         \caption{BBEH}
%         \label{fig: bbeh_alignment}
%     \end{subfigure}
    
%     \caption{The policy-reward reasoning manifold alignment under DCRL and static reward system settings.}
%     \label{fig: q2_alignment}
% \end{figure}

\subsection{Ablation Study}
\label{sec: ablation_study}

\begin{figure*}[t]
	\centering
  \includegraphics[width=\textwidth]{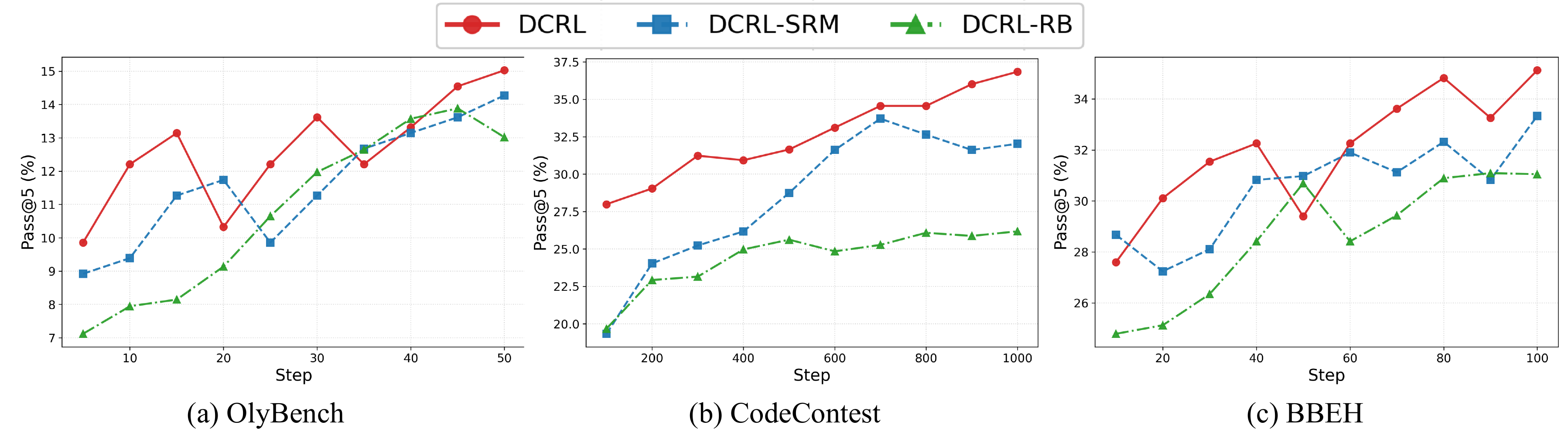}
  \caption{
    The results of the ablation experiment for DCRL. 
}
\label{fig: q3}
\end{figure*}

To address \textbf{Q3}, we conduct a comprehensive ablation study of DCRL across three representative reasoning domains: OlyBench for mathematical reasoning, CodeContest for code reasoning, and BBEH for commonsense reasoning. Specifically, as shown in Fig.~\ref{fig: q3}, the red curve labeled \textit{DCRL} denotes the validation performance of the full DCRL framework throughout the training process. The blue curve, \textit{DCRL-SRM}, represents the performance of DCRL when the syllogistic logic-based prompt evolution mechanism is replaced by a static reward model, while the green curve, \textit{DCRL-RB}, corresponds to the variant in which rewards are generated using a rule-based reward system. It can be observed that the full DCRL framework consistently outperforms the two ablated variants across all three datasets, highlighting the importance of mitigating the mismatch between the reasoning manifolds of the policy model and the reward system illustrated in Section~\ref{sec:Intro} and Section~\ref{sec: Metho}. In particular, Fig.~\ref{fig: q3}(a) and Fig.~\ref{fig: q3}(b) show that DCRL maintains a steady upward performance trend throughout training, indicating continuous improvements in reasoning capability. In contrast, the performance gains of both \textit{DCRL-SRM} and \textit{DCRL-RB} gradually plateau, with their improvement rates slowing significantly as training progresses. This divergence suggests that the decoupling and coupling mechanism in DCRL provides stronger scalability and sustained optimization benefits.

\section{Conclusion}
In this work, we introduced a geometric perspective on LLM reasoning by modeling it as a coupled manifold of logical deduction, evaluation, and representation, and identified the limitations of existing RL reward systems as a manifestation of policy–reward manifold mismatch. Building on this insight, we proposed the DCRL framework, which integrates dynamic reward rubric evolution with policy–reward re-coupling to achieve more stable and aligned optimization. Notably, DCRL enables Qwen3-4B to outperform its 32B counterpart and approach the 235B model on the evaluated benchmarks, demonstrating the effectiveness of policy–reward co-evolution. Both theoretical analysis and empirical results across diverse reasoning tasks demonstrate that DCRL consistently improves performance and generalization over conventional rule-based and reward-model approaches. These findings suggest that explicitly modeling and maintaining the structural alignment between policy and reward manifolds offers a promising direction for advancing RL-based post-training of LLMs.

%%%%%%%%%%%%%%%%%%%%%%%%%%%%%%%%%%%%%%%%%%%%%%%%%%%%%%%%%%%%

\newpage
\bibliography{custom}
\bibliographystyle{iclr2027_conference}

\newpage
\appendix

\section{Theoretical Proofs about DCRL}
\label{appendix: theo_proof}

\subsection{Theoretical Proof about Theorems~\ref{theo: bounded}}
To prove Theorem~\ref{theo: bounded}, we show that the prompt rewriting operator \(g_\phi\) can only induce a bounded perturbation on the prompt representation in each evolution step. More specifically, let the corollary prompt \(\mathcal C_t\) at epoch \(t\) consist of a sequence of \(n_t\) prompt tokens:
\begin{equation}
\mathcal C_t=\{w_1,w_2,\dots,w_{n_t}\}.
\label{eq:prompt_tokens}
\end{equation}

Each token \(w_i\) is mapped by the reward model embedding layer to a vector:
\begin{equation}
e_i\in\mathbb R^d,
\label{eq:token_embedding}
\end{equation}
where \(d\) denotes the embedding dimension. The full prompt representation is defined as the concatenation of all token embeddings:
\begin{equation}
E(\mathcal C_t)=\{e_1,e_2,\dots,e_{n_t}\}.
\label{eq:prompt_embedding}
\end{equation}

Thus, the prompt evolution operator
$g_\phi(\mathcal C_t,\bar A_t,\mathrm{Acc}_t)$
can be equivalently viewed as an operator acting on the prompt embedding:
\begin{equation}
E(\mathcal C_{t+1})
=
g_\phi\bigl(E(\mathcal C_t),\bar A_t,\mathrm{Acc}_t\bigr).
\label{eq:embedding_operator}
\end{equation}

Our goal is to bound the change:
\begin{equation}
\|
E(\mathcal C_{t+1})-E(\mathcal C_t)
\|.
\label{eq:goal_bound}
\end{equation}

At each epoch, the prompt rewriting operator updates the corollary prompt by rewriting part of the prompt tokens.
Assume that at most \(m\) tokens are modified during one prompt evolution step, where
\begin{equation}
m\le n_t.
\label{eq:m_bound}
\end{equation}

Let the modified token indices be
$i_1,i_2,\dots,i_m.$
For each modified token, the original embedding is \(e_{i_j}\), and the updated embedding is \(e'_{i_j}\). Then the prompt embedding difference can be written as the $L_1$ norm:
\begin{equation}
E(\mathcal C_{t+1})-E(\mathcal C_t)
=
\sum_{j=1}^{m}(e'_{i_j}-e_{i_j}).
\label{eq:embedding_difference}
\end{equation}

This equation means that the total prompt change is exactly the sum of the embedding changes of all modified tokens.
Due to the nature of the trust region in most of RL mechanisms nowadays, every token embedding produced by the reward model can be bounded in norm:
\begin{equation}
\|e_i\|\le M,
\qquad
\|e'_i\|\le M,
\label{eq:token_norm_bound}
\end{equation}
for some constant \(M>0\). For each modified token, by the triangle inequality:
\begin{equation}
\|e'_{i_j}-e_{i_j}\|
\le
\|e'_{i_j}\|+\|e_{i_j}\|.
\label{eq:triangle}
\end{equation}

Substituting Eq.~\eqref{eq:token_norm_bound} into Eq.~\eqref{eq:triangle}, we obtain:
\begin{equation}
\|e'_{i_j}-e_{i_j}\|
\le
2M.
\label{eq:single_token_bound}
\end{equation}

Therefore, each modified token contributes at most \(2M\) to the total prompt embedding change.
Then, taking norms on both sides of Eq.~\eqref{eq:embedding_difference} and applying the triangle inequality gives:
\begin{equation}
\|
E(\mathcal C_{t+1})-E(\mathcal C_t)
\|
\le
\sum_{j=1}^{m}
\|e'_{i_j}-e_{i_j}\|.
\label{eq:sum_bound}
\end{equation}

Using Eq.~\eqref{eq:single_token_bound}, we further obtain:
\begin{equation}
\|
E(\mathcal C_{t+1})-E(\mathcal C_t)
\|
\le
\sum_{j=1}^{m}2M
=
2mM.
\label{eq:total_bound}
\end{equation}

Thus, the prompt rewriting operator induces a bounded prompt perturbation with upper bound:
\begin{equation}
\delta=2mM.
\label{eq:delta_def}
\end{equation}

Substituting Eq.~\eqref{eq:delta_def} into Eq.~\eqref{eq:total_bound}, we obtain:
\begin{equation}
\|
E(\mathcal C_{t+1})-E(\mathcal C_t)
\|
\le
\delta.
\label{eq:bounded_prompt_result}
\end{equation}

Since the prompt rewriting operator \(g_\phi\) modifies at most \(m\) prompt tokens in one evolution step and each token embedding norm is bounded by \(M\), the induced prompt perturbation is bounded by the constant
\[
\delta=2mM.
\]

Therefore, the prompt evolution step satisfies
\begin{equation}
\|
\mathcal C_{t+1}-\mathcal C_t
\|
\le
\delta,
\label{eq:final_theorem_bound}
\end{equation}
which proves Theorem~\ref{theo: bounded}.

\subsection{Theoretical Proof about Theorems~\ref{theo: gap}}
In this appendix, we provide the complete proof that the reasoning-manifold gap between the policy model and reward model remains bounded under DCRL.
At training epoch \(t\), let
$Z_t^\pi\sim p_t^\pi(z)$
denote the latent reasoning representation induced by the policy model, and let
$Z_t^r\sim p_t^r(z)$
denote the latent reasoning representation induced by the reward model.

The corresponding reasoning manifolds are defined as the supports of these latent distributions:
\begin{equation}
\mathcal M_t^\pi
=
\mathrm{supp}(p_t^\pi),
\qquad
\mathcal M_t^r
=
\mathrm{supp}(p_t^r).
\label{eq:manifold_def}
\end{equation}

To quantify the discrepancy between the two reasoning manifolds, define the symmetric KL divergence:
\begin{equation}
G_t
=
D_{\mathrm{KL}}(p_t^\pi\|p_t^r)
+
D_{\mathrm{KL}}(p_t^r\|p_t^\pi).
\label{eq:gap_def}
\end{equation}

Our goal is to prove that \(G_t\) remains bounded by a constant during the DCRL optimization process.

Recall the parameter synchronization strategy from Eq.~(\ref{eq: policy_updation}) in Algorithm~\ref{alg: dcrl}, at the end of each epoch, the reward model parameters are synchronized with the policy model:
\begin{equation}
\phi_{t+1}
\leftarrow
\theta_{t+1}.
\label{eq:sync}
\end{equation}

This synchronization ensures that before prompt evolution, the reward model and the policy model share identical parameters.
Therefore, the latent reasoning distribution of the reward model before prompt rewriting is identical to that of the policy model:
\begin{equation}
\tilde p_{t+1}^r(z)
=
p_{t+1}^\pi(z).
\label{eq:aligned}
\end{equation}

This means that immediately after synchronization, the discrepancy between the policy reasoning manifold and the reward reasoning manifold is zero.
After synchronization, the corollary prompts are updated through the prompt rewriting operator:
\begin{equation}
\mathcal C_{t+1}
=
g_\phi(\mathcal C_t,\bar A_t,\mathrm{Acc}_t).
\label{eq:update_appendix}
\end{equation}

This prompt update changes the reward reasoning distribution through the prompt transformation operator \(T_{\mathcal C}\), producing:
\begin{equation}
p_{t+1}^r(z)
=
T_{\mathcal C_{t+1}}p_{t+1}^\pi(z).
\label{eq:transform}
\end{equation}

Hence, after synchronization, the only factor that can create discrepancy between the policy reasoning distribution and the reward reasoning distribution is the prompt evolution step.
By Theorem~\ref{theo: bounded}, the prompt rewriting operator satisfies:
\begin{equation}
\|
\mathcal C_{t+1}-\mathcal C_t
\|
\le
\delta.
\label{eq:delta_bound}
\end{equation}

This means that the change introduced to the reward prompts at each epoch is limited by the constant \(\delta\).
By Assumption~\ref{assump:smooth}, the KL divergence between two reward reasoning distributions induced by two nearby prompts is bounded by the prompt difference:
\begin{equation}
D_{\mathrm{KL}}
\bigl(
p_{t+1}^r
\|
p_{t+1}^\pi
\bigr)
\le
L
\|
\mathcal C_{t+1}-\mathcal C_t
\|.
\label{eq:forward_bound_1}
\end{equation}

Substituting Eq.~\eqref{eq:delta_bound} into Eq.~\eqref{eq:forward_bound_1}, we obtain:
\begin{equation}
D_{\mathrm{KL}}
\bigl(
p_{t+1}^r
\|
p_{t+1}^\pi
\bigr)
\le
L\delta.
\label{eq:forward_bound_2}
\end{equation}

Therefore, the divergence from the reward reasoning distribution to the policy reasoning distribution is bounded by \(L\delta\).
Since the prompt perturbation is bounded and smooth, the reverse KL divergence can be bounded analogously:
\begin{equation}
D_{\mathrm{KL}}
\bigl(
p_{t+1}^\pi
\|
p_{t+1}^r
\bigr)
\le
L\delta.
\label{eq:reverse_bound}
\end{equation}

Thus, both directions of the manifold discrepancy are bounded by the same constant.
Substituting Eqs.~\eqref{eq:forward_bound_2} and~\eqref{eq:reverse_bound} into Eq.~\eqref{eq:gap_def}, we obtain:
\begin{align}
G_{t+1}
&=
D_{\mathrm{KL}}(p_{t+1}^\pi\|p_{t+1}^r)
+
D_{\mathrm{KL}}(p_{t+1}^r\|p_{t+1}^\pi)
\notag\\
&\le
L\delta+L\delta
\notag\\
&=
2L\delta.
\label{eq:gap_bound}
\end{align}

Therefore,
\begin{equation}
G_t\le 2L\delta,\qquad \forall t.
\label{eq:final_result}
\end{equation}

This proves that the discrepancy between the policy reasoning manifold and the reward reasoning manifold remains bounded during the entire DCRL training process.

\section{Datasets and Experiments Pre-processing}
\label{appen: pre-processing}

\subsection{Dataset Pre-processing}
\label{appen: dataset_pre_processing}

This section provides an overview of the datasets employed in our reinforcement learning pipeline and describes the corresponding preprocessing and normalization procedures implemented in our experiments.

The pre-processing framework is designed to preserve the native supervision signals of heterogeneous benchmarks while minimizing noise introduced by output formatting. First, the unified data schema enables a single reinforcement learning framework to process diverse task types. Second, dataset-specific normalization ensures that reward signals reflect semantic correctness rather than superficial output variations. Third, backend fallback mechanisms and prewarm validation improve reproducibility across restricted execution environments. Finally, prompt-level formatting constraints, particularly for mathematical reasoning datasets, improve the robustness of answer extraction. Collectively, these design choices stabilize the cross-dataset reinforcement learning pipeline while preserving the semantic integrity of each benchmark.

To ensure compatibility across heterogeneous benchmarks, all datasets are standardized into a unified chat-style representation accompanied by structured supervision metadata. Specifically, the \texttt{prompt} field contains a sequence of chat messages, typically including a user query and an auxiliary system instruction. 
To maintain consistency during the experiments, identical dataloader constraints are applied to all datasets, including \texttt{max\_prompt\_length=6000}, \texttt{max\_response\_length=6000}, \texttt{filter\_overlong\_prompts=True}, \texttt{truncation=error}, and random shuffling. The detailed illustration about the benchmarks employed in the experiments is presented below.

\medskip
\noindent\textbf{GSM8K.}
GSM8K~\citep{cobbe2021training} is a grade-school mathematical word-problem benchmark consisting of approximately 8.5K instances, designed to evaluate multi-step arithmetic reasoning. In our framework, GSM8K is treated as a mathematical reasoning dataset with numeric-answer supervision. During preprocessing and evaluation, commas and currency symbols are removed from both predictions and labels. 

\medskip
\noindent\textbf{AIME24 and AIME25.}
AIME24 and AIME25~\citep{balunovic2025matharena} are olympiad-style mathematical reasoning benchmarks derived from the American Invitational Mathematics Examination, where answers are integers in the range \([0,999]\). In our implementation, these datasets are evaluated using an exact symbolic matching verifier with normalization. Any serialized list-like outputs are transformed into canonical scalar forms before evaluation to ensure consistency.

\medskip
\noindent\textbf{OlyBench.}
OlyBench~\citep{he2024olympiadbench} serves as an olympiad-level mathematical reasoning benchmark in our experiments. To facilitate reliable reward extraction, we append an explicit instruction requiring the model to produce the final answer in the format \texttt{Final Answer: <number>}. The processed prompts are cached for efficiency and automatically regenerated whenever the source data or formatting instruction changes.

\medskip
\noindent\textbf{CodeContest.}
CodeContest~\citep{li2022competition} is derived from DeepMind's competitive programming benchmark \textit{CodeContests}, containing programming problems paired with test cases and problem metadata. Prior to training, a prewarm verification step retrieves a sample problem to ensure that external test-case access functions correctly. During evaluation, public, private, and generated tests are executed under bounded time limits to maintain stable training throughput.

\medskip
\noindent\textbf{L.C.Bench.}
L.C.Bench (i.e., LiveCodeBench)~\citep{jain2024livecodebench} is a contamination-aware code generation benchmark composed of recent contest problems from platforms such as LeetCode, AtCoder, and Codeforces. During preprocessing, each problem is converted into the unified prompt format, while the corresponding \texttt{question\_id} is stored for test-case retrieval. The resulting dataset is deterministically partitioned into training and validation sets. During reward evaluation, the generated solutions are executed against sandboxed unit tests associated with each \texttt{question\_id}.

\medskip
\noindent\textbf{BBEH.}
BBEH (i.e., BIG-Bench Extra Hard)~\citep{kazemi2025big} extends the BBH benchmark with more challenging tasks while preserving broad reasoning coverage.  Each task is transformed into the unified schema with explicit final-answer prompting, and answer candidates are normalized into canonical forms (e.g., option letters, yes/no responses, or numeric strings) before exact-match evaluation.

\medskip
\noindent\textbf{MMLU-Pro.}
MMLU-Pro~\citep{wang2024mmlu} is an enhanced version of MMLU featuring more difficult reasoning tasks and ten-way multiple-choice options, thereby reducing benchmark saturation. In our preprocessing pipeline, both labels and model outputs undergo option normalization, prioritizing explicit final-answer patterns and falling back to option extraction heuristics. The reward signal is determined by exact agreement between normalized option labels.

\medskip
\noindent\textbf{GPQA.}
GPQA~\citep{rein2023gpqa} is a graduate-level multiple-choice reasoning benchmark covering advanced scientific domains such as biology, chemistry, and physics. During preprocessing, incomplete entries are removed, distractor options are randomized, and answer labels are balanced to ensure near-uniform distribution across options. Prompts are then formatted with explicit final-answer instructions before being exported to parquet files.

\subsection{Experiments Pre-processing in Section~\ref{sec: manifolds_alignment}}
\label{appen: experiment_pre_processing}

To quantitatively examine the alignment dynamics between the policy model and the reward model during training, we conduct a step-wise manifold alignment analysis comparing the \texttt{coevolve} and \texttt{non\_coevolve} training settings. The objective of this experiment is to measure how closely the representation manifolds induced by the policy outputs and reward-model outputs evolve together throughout the proposed DCRL optimization process. Specifically, we adopt two complementary metrics---\textbf{Linear Centered Kernel Alignment (CKA)} and \textbf{Procrustes Distance}---to characterize the structural similarity and geometric discrepancy between the two manifolds.

\paragraph{Data Collection.}
For each training configuration, rollout records are collected from the corresponding JSONL logs generated during policy optimization. Each valid rollout record contains the policy response text, the reward-model response text, and the associated optimization indices, including the training step and epoch number. Formally, each record can be represented as:
\[
(x_i^{(t)}, y_i^{(t)}, t, e),
\]
where $x_i^{(t)}$ denotes the policy output, $y_i^{(t)}$ denotes the reward-model output, $t$ is the optimization step, and $e$ is the epoch index. These records are grouped by task family and then partitioned into two experimental sets corresponding to the co-evolutionary and non-co-evolutionary reward training paradigms:
\[
\mathcal{D}^{\mathrm{co}}, \qquad \mathcal{D}^{\mathrm{non}}.
\]

\paragraph{Text Representation Construction.}
To compare the latent structures of policy outputs and reward-model outputs, both textual responses are transformed into fixed-dimensional vector representations. We employ a two-stage embedding pipeline. First, textual responses are encoded using TF-IDF features, followed by truncated singular value decomposition (SVD) to obtain dense semantic representations. If TF-IDF features are unavailable due to sparsity constraints, a hashed bag-of-tokens embedding is used as a fallback strategy.

For each paired sample at training step $t$, this procedure yields a policy embedding vector $\mathbf{p}_i^{(t)} \in \mathbb{R}^d$ and a reward embedding vector $\mathbf{r}_i^{(t)} \in \mathbb{R}^d$. Aggregating all $n_t$ valid samples at step $t$ forms the policy and reward embedding matrices:
\[
P_t =
\begin{bmatrix}
(\mathbf{p}_1^{(t)})^\top\\
\vdots\\
(\mathbf{p}_{n_t}^{(t)})^\top
\end{bmatrix}
\in \mathbb{R}^{n_t\times d},
\qquad
R_t =
\begin{bmatrix}
(\mathbf{r}_1^{(t)})^\top\\
\vdots\\
(\mathbf{r}_{n_t}^{(t)})^\top
\end{bmatrix}
\in \mathbb{R}^{n_t\times d}.
\]

Although the underlying policy and reward reasoning manifolds are not directly observable, their generated textual outputs provide observable projections of the corresponding latent reasoning processes. At training step \(t\), the policy model maps each input query to a response, while the reward model maps the same policy behavior into an evaluative response or judgment. We therefore regard the collections of policy-output embeddings \(P_t\) and reward-output embeddings \(R_t\) as empirical, output-space approximations of the policy and reward reasoning manifolds, respectively. In this view, each row of \(P_t\) or \(R_t\) represents one sampled point on the observable projection of the corresponding manifold, and the geometry induced by pairwise relations among these samples reflects how the two models organize reasoning behavior at that step. Consequently, Linear CKA and Procrustes Distance do not measure the full latent manifolds directly, but quantify the alignment between their empirical textual projections, which serves as a practical proxy for analyzing policy-reward manifold alignment during training.

\paragraph{Linear CKA for Structural Similarity.}
To measure the structural similarity between the policy and reward manifolds, we compute the Linear CKA score at each training step. We first center the feature matrices:
\[
\tilde{P}_t = P_t - \mathbf{1}\mu_{P_t}^\top,\qquad
\tilde{R}_t = R_t - \mathbf{1}\mu_{R_t}^\top,
\]
where $\mu_{P_t}$ and $\mu_{R_t}$ denote the column-wise means.

The step-wise Linear CKA score is then defined as:
\[
\mathrm{CKA}_t =
\frac{\left\|\tilde{P}_t^\top\tilde{R}_t\right\|_F^2}
{\sqrt{
\left\|\tilde{P}_t^\top\tilde{P}_t\right\|_F^2
\left\|\tilde{R}_t^\top\tilde{R}_t\right\|_F^2
}}.
\]

A larger $\mathrm{CKA}_t$ indicates that the policy and reward representations share stronger structural similarity, suggesting that the reward manifold evolves in a manner that better tracks the policy manifold during training.

\paragraph{Procrustes Distance for Geometric Alignment.}
To further evaluate the geometric discrepancy between the two manifolds, we compute the orthogonal Procrustes distance at each step. After centering, the matrices are normalized by their Frobenius norms:
\[
\hat{P}_t=\frac{\tilde{P}_t}{\|\tilde{P}_t\|_F},
\qquad
\hat{R}_t=\frac{\tilde{R}_t}{\|\tilde{R}_t\|_F}.
\]

We then solve the orthogonal alignment problem:
\[
M_t = \hat{P}_t^\top\hat{R}_t = U_t\Sigma_tV_t^\top,
\qquad
Q_t = U_tV_t^\top,
\]
and define the normalized Procrustes distance as:
\[
\mathrm{Proc}_t =
\frac{\|\hat{P}_tQ_t-\hat{R}_t\|_F}{\sqrt{n_t}}.
\]

This metric measures the residual discrepancy between the two manifolds after optimal orthogonal transformation. A smaller $\mathrm{Proc}_t$ implies better geometric consistency between the policy and reward manifolds.

\paragraph{Step-wise Metric Aggregation.}
For each experimental setting, the two alignment metrics are computed at every valid optimization step:
\[
\{(\mathrm{CKA}_t,\mathrm{Proc}_t)\}_{t\in\mathcal{T}}.
\]
To enable comparison across runs with different absolute step counts, all metrics are plotted against normalized training progress ranging from $0\%$ to $100\%$. For visualization purposes, a moving average smoothing operation may be applied, while the raw step-level values are retained for statistical analysis.

\paragraph{Summary Statistics.}
To summarize the overall alignment behavior over the full training trajectory, we compute the mean CKA and mean Procrustes distance:
\[
\overline{\mathrm{CKA}}
=
\frac{1}{|\mathcal{T}|}\sum_{t\in\mathcal{T}}\mathrm{CKA}_t,
\qquad
\overline{\mathrm{Proc}}
=
\frac{1}{|\mathcal{T}|}\sum_{t\in\mathcal{T}}\mathrm{Proc}_t.
\]

The relative advantage of co-evolutionary reward learning over the non-co-evolutionary baseline is measured by:
\[
\Delta\overline{\mathrm{CKA}}
=
\overline{\mathrm{CKA}}^{\mathrm{co}}
-
\overline{\mathrm{CKA}}^{\mathrm{non}},
\qquad
\Delta\overline{\mathrm{Proc}}
=
\overline{\mathrm{Proc}}^{\mathrm{co}}
-
\overline{\mathrm{Proc}}^{\mathrm{non}}.
\]

\section{Example Prompts for DCRL}
The prompts below are examples for formation curation.

\begin{promptbox}
\begin{PromptVerbatim}
Answer using exactly one final line in this format: Final Answer: <OPTION>
Where <OPTION> must be one of: A, B, C, D, Yes, No.
Example: Final Answer: A
\end{PromptVerbatim}
\end{promptbox}

\begin{promptbox}
\begin{PromptVerbatim}
Return the final result on the last line exactly as:
Final Answer: <number>.
Do not add any text after that line.
\end{PromptVerbatim}
\end{promptbox}

The prompts below are examples of syllogistic logic system prompt.

\begin{promptbox}
\begin{PromptVerbatim}
You are an expert reward model judge for reinforcement-learning reward shaping.
Follow the Aristotle three-paradigm hierarchy strictly:

[Axioms - immutable and cross-domains]
- Axiom 1: Functional correctness has highest priority over style, verbosity, or formatting.
- Axiom 2: Reward must be grounded in the given problem statement and observable solution behavior.
- Axiom 3: Do not use length, position, or model-identity biases when scoring.
- Axiom 4: Penalize hallucinated assumptions and unverifiable claims.

[Theorems - updatable medium-level rules when reasoning domain is changed]
- (none)  # dynamically replaced when changing the reasoning domain

[Corollaries - updatable tactical heuristics]
- (none)  # dynamically replaced by runtime

Execution-based evaluation is the primary signal; your score is an auxiliary reward.
\end{PromptVerbatim}
\end{promptbox}

% \begin{promptbox}
% \begin{PromptVerbatim}
% Problem:
% {problem}

% Candidate Reasoning (optional):
% {thinking}

% Candidate Code:
% {solution}

% Evaluation procedure:
% 1) Identify the required task from the problem.
% 2) Check whether the code logic matches the task.
% 3) Assess correctness on normal and edge cases.
% 4) Assess robustness and efficiency only after correctness.
% 5) Assign exactly one integer score from 1 to 10 using the rubric.

% Scoring rubric (single integer 1-10):
% 1 = Empty/irrelevant output.
% 2 = Fundamental misunderstanding; approach cannot solve the stated task.
% 3 = Clearly incorrect logic; fails almost all meaningful cases.
% 4 = Major algorithmic or data-flow errors; only superficial alignment with task.
% 5 = Partially correct; solves limited/simple cases but misses key logic.
% 6 = Mostly correct; still has important bug(s) or misses notable edge cases.
% 7 = Correct on typical cases; minor robustness or completeness gaps.
% 8 = Correct and clean; handles most edge cases with small remaining weaknesses.
% 9 = Correct, robust, and well-structured; good edge-case coverage and sound choices.
% 10 = Fully correct and robust with strong clarity.

% Output rule:
% Reply with only one integer from 1 to 10, and nothing else.
% \end{PromptVerbatim}
% \end{promptbox}

\begin{promptbox}
\begin{PromptVerbatim}
You are a reward model performing decoupling and coupling RL framework for better scoring quality.
You must keep axioms immutable and only update theorem/corollary level rules.

Current prompt state:
{json_current_state}

Recent scoring experience samples:
{json_experience_samples}

Task:
1) Derive scoring experience that can improve accuracy.
2) Keep axioms unchanged.
3) Propose updated theorem and corollary rules.
4) Theorem can only be changed when reasoning domain is shifted; corollary should be tactical heuristics.
5) Return STRICT JSON object with keys:
   {"theorems": [..], "corollaries": [..], "rationale": "..."}
Do not include markdown fences.
\end{PromptVerbatim}
\end{promptbox}

\section{Limitations}
\label{appen: limitations}
In this section, we critically analyze the limitations of the proposed DCRL framework, which can be summarized along two primary dimensions: its applicability to agentic RL and its computational overhead.
First, regarding applicability to agentic RL, the current formulation of DCRL is primarily designed for RL in the context of LLMs. However, with the increasing prominence of agent-based systems, agentic RL has emerged as a crucial paradigm. Extending DCRL to such settings is non-trivial, as it remains unclear how the proposed framework can be effectively adapted to accommodate the dynamic, multi-step decision-making processes inherent in agentic environments.
Second, in terms of computational overhead, the syllogistic prompt-evolving mechanism introduced in this work incurs additional computational costs. Moreover, its performance may be sensitive to the design of prompts and the stability of their updates, potentially affecting the robustness and efficiency of the overall training process.

\section*{AI Use Statement}
% (This section is \textbf{required} and does not count toward the page limit.)
In this work, we used generative AI tool to assist with translation, implement methods with codes and provide feedback on experiments.
We have not used generative AI tools for generate synthetic data sets, help develop conceptual frameworks, formulate mathematical claims, provide critical ingredients for proving mathematical claims, assist in the writing of proofs, propose hypotheses, clean the dataset, support qualitative and thematic data analysis and interpret results. We have reviewed all AI-assisted work. For example, LLM-generated code was verified and tested for correctness
by 2 authors. We take responsibility for the final content of this work, including text, claims or artifacts produced with the aid of generative AI.

%%%%%%%%%%%%%%%%%%%%%%%%%%%%%%%%%%%%%%%%%%%%%%%%%%%%%%%%%%%%

\end{document}